\documentclass[10pt,twocolumn,letterpaper]{article}

\usepackage[pagenumbers]{cvpr} 
\makeatletter
\@namedef{ver@everyshi.sty}{}
\makeatother
\usepackage{tikz}
\usepackage{graphicx}
\usepackage{amsmath}
\usepackage{amssymb}
\usepackage{booktabs}
\usepackage{makecell}

\usepackage{todonotes}

\usepackage[pagebackref,breaklinks,colorlinks]{hyperref}

\usepackage[capitalize]{cleveref}
\crefname{section}{Sec.}{Secs.}
\Crefname{section}{Section}{Sections}
\Crefname{table}{Table}{Tables}
\crefname{table}{Tab.}{Tabs.}

\def\cvprPaperID{5} 
\def\confName{CVPR}
\def\confYear{2022}

\begin{document}

\title{Leveraging Unlabeled Data for Sketch-based Understanding}

\author{Javier Morales\\
Department of Computer Science\\
University of Chile, Chile \\
{\tt\small javiermoralesr95@gmail.com}
\and
Nils Murrugarra-Llerena\\
Department of Computer Science\\
Weber State University\\
{\tt\small nmurrugarrallerena@weber.edu}
\and
Jose M. Saavedra\\
Universidad de los Andes, Chile\\
{\tt\small jmsaavedrar@miuandes.cl}
}
\maketitle

\begin{abstract}
Sketch-based understanding is a critical component of human cognitive learning and is a primitive communication means between humans. This topic has recently attracted the interest of the computer vision community as sketching represents a powerful tool to express static objects and dynamic scenes. Unfortunately, despite its broad application domains, the current sketch-based models strongly rely on labels for supervised training, ignoring knowledge from unlabeled data, thus limiting the underlying generalization and the applicability. Therefore, we present a study about the use of unlabeled data to improve a sketch-based model. To this end,  we evaluate variations of VAE and semi-supervised VAE, and present an extension of BYOL to deal with sketches. Our results show the superiority of sketch-BYOL, which outperforms other self-supervised approaches increasing the retrieval performance for known and unknown categories. Furthermore, we show how other tasks can benefit from our proposal.
\end{abstract}


\section{Introduction}
Sketch-based understanding plays an important role in the visual perception system. During the half last century, Hubel and Wiesel \cite{book:hubel_wiesel} had already shown how the biological visual cortex highly responds to edge patterns. More recently,  Walther et al. \cite{walther:2011} also showed the semantic power of image contour information through a study of functional Magnetic Resonance Imaging (fMRI). They found that the primary visual cortex produces similar responses when stimulated by a regular image or by its corresponding contour map.  

Sketch understanding is deeply connected to cognition development \cite{forbus:2011}. Sketching is the means through which an infant starts to understand the natural environment, and also it enables people to externalize and communicate simple and complex ideas. Indeed, people draw schemes or maps to understand complex structures and unfold complex processes. In this vein, Mukherjee et al. \cite{mukherjee:2019} studied how we effortlessly associate a drawing with objects in the world. The authors found that the compositional nature of object concepts allows us to decompose objects and drawings into semantically meaningful parts. 

Due to the critical role that sketch-based understanding plays in the visual perception process, together with the ubiquitous use of touch-screen devices that make sketching a convenient mechanism, the computer vision community has started to pay special attention to this area. For instance, the main computer vision conferences already include workshops to promote research and applications on this topic. In this vein,  we have seen advances in a diversity of tasks like sketch classification \cite{eitz:2012, yu:2017, Xu_2018_CVPR},  sketch-guided object localization \cite{tripathi:2020}, sketch-based image and video retrieval \cite{bui:2018, torres:2021, fuentes:2021, murrugarra_2018_bmvc, murrugarra_2021_cviu, collomosse:2009, Yu:2021}, sketch-to-photo translation \cite{chen:2018, sangkloy:2017}, among others.

However, as far as we know, the sketch-based models strongly rely on labeled data \cite{Xu_2018_CVPR, deep_tpami_2022}. These models need sketches to be annotated with their classes or connected with corresponding images (making pairs)  to train supervised models. Having this strong dependence on labeled data raises three critical problems: i) it limits the applicability as labeling is an impractical task for industry, ii) it wastes a vast amount of unlabeled data, and iii) it limits the generalization of learned representations.

This work aims to tackle these limitations by leveraging unlabeled data and creating accurate representations from sketches. We study self-supervised approaches like VAE \cite{kingma2014autoencoding} and BYOL \cite{NEURIPS2020_f3ada80d}), semi-supervised approaches and traditional supervised models \cite{he:2016}  for sketch retrieval. Our semi-supervised VAE baselines adapt VAE, and add a classification branch via sampling concatenation or classification loss. Also, we extend BYOL to work in the sketch domain.

We compare the proposed approaches under known and unknown categories. For known categories, the best performer is a supervised model (ResNet-50), as expected. However, it does not generalize well for unknown categories. While sketch-BYOL shows competitive performance for known categories, and is the best performer for unknown categories, showing a better generalization power. This finding was confirmed by embedding visualization and sketch retrieval examples. BYOL better differentiates categories, and shows more intuitive retrievals. Furthermore, we present an example of the utility of our approach to allow self-supervision in other tasks where making sketch-image pairs is a critical stage like sketch-to-photo generation, sketch-based localization, and sketch-based image retrieval.

In summary, our main contributions are:

\begin{itemize}
    \item A study of multiple ways of mine unlabeled data to improve sketch understanding. This study considers semi-supervised and self-supervised approaches. For self-supervised models, we propose sketch-BYOL discovering which transformations are effective for sketch understanding.
    \item An strategy to allow self-supervision in tasks where making sketch-image pairs is critical. 
\end{itemize}
\section{Related work}
Sketching is a new emerging modality with its characteristics and challenges. Sketches can communicate abstract ideas from humans to machines \cite{murrugarra_2018_bmvc, murrugarra_2021_cviu}, and they are subject to different human drawing styles \cite{Xu2021DeepSR}. Also, as opposed to a static image pixel representation, sketches can be modeled as a temporal stroke sequence \cite{Xu_2018_CVPR, ha_2018_iclr}, and also as topological representations via graphs \cite{xu2021multigraph, deep_tpami_2022}.

Related to improving sketch-based retrieval,\cite{Xu_2018_CVPR} develop a novel sketch hashing retrieval technique and a CNN\--RNN network to understand millions of sketches accommodating their large variations in styles and abstractions. Similarly to combining CNN and RNN, \cite{Xu2021DeepSR} combines textual convolutional network with CNNs to create a self-supervised representation for sketches. Their main contribution is a set of geometric deformation to create variability and diversity in sketches, and they serve as pretext tasks for self-supervised learning. Also, from unsupervised learning, \cite{zhang_unsup_idetc_2020} learn a latent space to group different ``visual prototypes" using a clustering layer.

Our work complements these efforts, and similarly to \cite{Xu2021DeepSR} uses self-supervised learning. Similarly, we identify sketch transformations such as rotation, line skip, flip, and crop under a BYOL framework \cite{NEURIPS2020_f3ada80d}.

\subsection{Sketch-based classification}
Image classification is the most popular task in computer vision. A diversity of models have been proposed for classification \cite{eitz:2012, yu:2017} or learning representation from sketches \cite{Xu_2018_CVPR, ha_2018_iclr, Xu:2021-1} achieving high accuracy. These advances were achieved due to the availability of sketch datasets like QuickDraw or Sketchy \cite{sangkloy:2017}. 

Although we have seen good results in public datasets, we have a critical limitation in industry application. The models rely on a huge amount of labeled data, which is scarce or impractical in applications like e-commerce search engines. In this work, we propose to leverage unlabeled sketches to improve retrieval power, especially for unseen categories.


\subsubsection{Sketch-based image retrieval}
Sketch-based image retrieval (SBIR) is a growing field
in computer vision that consists of retrieving a collection
of photos resembling a sketched query. Aiming to make
the querying process as easy as possible, the input query is
formulated as a simple hand-drawing, composed uniquely
of strokes. Recent works in this task include that of  Bui et al. \cite{bui:2018}, proposing an incremental training process based on siamese networks; the work of Torres and Saavedra \cite{torres:2021} that showed the effectiveness of learning low-dimensional embedding using a local-topology preserving dimensional reduction  \cite{mcinnes:2018}.  A natural extension of SBIR is the case where the input sketch includes color information. Here, Fuentes and Saavedra \cite{fuentes:2021} recently presented an interesting approach extending the notion of triplets to quadruplet-based training.

As opposed to these related work, we deal with sketch retrieval under semi and self-supervised learning. We also show how our results are applied to increase the variability in making sketch-image pairs for training a sketch-based image retrieval model.


\subsubsection{Sketch-based localization}
The idea for a model is to localize all instances of an object in a regular image (scene). A sketch represents the target object. The model should respond with a bounding box enclosing the target object. In this context, Tripathi et al. \cite{tripathi:2020} combines a siamese network, cross-attention, and a region proposal model to train a generalized sketch-based localization model.

Our results yielded by our proposal sketch-BYOL can also be applied to support this task, as we can add variability to the query sketch during the training stage.

\subsubsection{Sketch-to-photo translation}

Sketch-to-photo translation 
aims to produce a photorealistic image from an input sketch. 
Researchers have proposed a diversity of approaches to deal with this problem \cite{chen:2018, sangkloy:2017,zhu:2017, isola:2017}, but to produce plausible results, they strongly depend on labeled sketch-photo pairs.

Our proposal can also fuel this task by generating sketch-photo pairs, in a self-supervised regimen, from edge maps to hand-drawn sketches.

\subsubsection{Sketch-based video retrieval}
Sketching is a powerful tool for representing static objects and dynamic scenes like videos. If an image is worth more than a thousand words, a sketch may be worth more than multiple images. For instance, a simple drawing representing a person with a left arrow can express the situation when someone moves in the right direction. Thus, sketch-based video understanding is another attractive task in this domain. Here, Collomosse et al. \cite{collomosse:2009} introduced sketches for content-based video retrieval. More recently,  Xu et al. \cite{Yu:2021} proposed a convnet-based model for fine-grained video retrieval, combining appearance and motion information with a relation module between sketch-video pairs.

As we can see, the last years have been marked by significant advances in the development of models or architectures addressing diverse problems based on sketch understanding 
However, the discussed advances share a common weakness. All of them depend on a huge amount of labeled data, which sets a limitation in real-world applications.

Therefore in this work, we explore and evaluate a diverse set of self and semi-supervised models in the sketch domain. We evaluate generative models like VAE \cite{kingma2014autoencoding} and discriminative models like BYOL \cite{NEURIPS2020_f3ada80d}. We evaluate our results in terms of how well our models generalize to unknown objects, and particularly to unknown classes.

Furthermore, building image and sketch pairs is a traditional annotation process for the tasks above, trained under a supervised learning strategy. This process places a challenge on a self-supervised strategy. We will show that our proposal is an efficient and effective way to deal with this challenge. Having an image, we could start with its corresponding contour map and search for similar human-drawn sketches to add variability to the initial contour. This could also be regarded as a sketch-based augmentation. 

\subsection{Self-supervised learning}
Self-supervision was mainly related to reconstruction-based generative models like Variational Autoencoder (VAE)\cite{kingma2014autoencoding}. It receives an input and encodes it to a low-dimensional vector, then decodes that vector to reconstruct the same input. 
VAE encoder produces two vectors a $\mu$ and a $log\; \sigma^2$, that together define a conditional probability distribution given the input.

However, more recently, we have seen high effectiveness of discriminative models. Here,  Grill et al. \cite{NEURIPS2020_f3ada80d} proposed BYOL achieving high performance on image representation learning. It comprises two networks, an online network, and a target network. BYOL is fed by two views from the same input image applying two different image transformations. It is then trained so that both networks produce the same latent vectors. 

Therefore, inspired by BYOL, we propose sketch-BYOL working in the sketch domain, identifying specific transformations for increasing accuracy.

\subsection{Semi-supervised learning}
M2 \cite{kingma2014semi} combines 
the output of an encoder with a label for reconstruction. The model uses the real label from the labeled data and the predicted label from the unlabeled data. A variation of this process is the \textbf{Y} shaped model, where a classifier is trained only with the labeled data.

We adapt and evaluate these two approaches for the sketch domain. Additional details come in the next section.
\section{Approach}
\subsection{Self-supervised approaches}

\subsubsection{Variational Autoencoder (VAE)}

\begin{figure}[ht!]
    \centering
    \includegraphics[width=\linewidth]{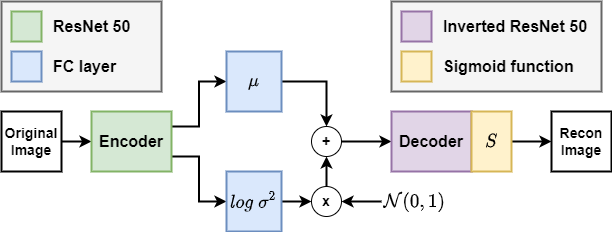}
    \caption{Proposed architecture for the VAE model. the encoder consists of a ResNet50 model and the latent space vectors are extracted with fully-connected layers. The decoder is an Inverted ResNet50 with a sigmoid activation function due to binary representation for sketches.}
    \label{fig:vae}
\end{figure}

The proposed architecture is shown in Figure \ref{fig:vae}, we utilize a ResNet50 \cite{he:2016}  as the encoder and an inverted ResNet50 for the decoder. We also use two fully-connected layers to extract 
latent vectors ($\mu$ and $log\; \sigma^2$) from the encoder. The model considers sketches of size $256\times 256$.

For sketch retrieval purposes, we utilize $\mu$ vectors with size of 32\footnote{From preliminary experiments, this configuration achieves the best performance}. 

We represent sketches with strokes (value 0) and background (value 1). Then, we simplify the decoder output with a sigmoid function and use a pixel-wise binary cross-entropy loss for reconstruction. We also use the  KL divergence (KLD) loss to better distribute the categories. KDL is weighted by $\beta = 0.1$ \cite{Higgins2017betaVAELB}, which showed to improve the results. Thus, the VAE unlabeled loss $u\mathcal{L}_{VAE}$ is defined in Equation \ref{eq:var_autoencoder}.

\begin{equation}
u\mathcal{L}_{VAE} = reconstruction + \beta KLD    
\label{eq:var_autoencoder}
\end{equation}

\subsubsection{Sketch-BYOL}
\begin{figure}[ht!]
    \centering
    \includegraphics[width=\linewidth]{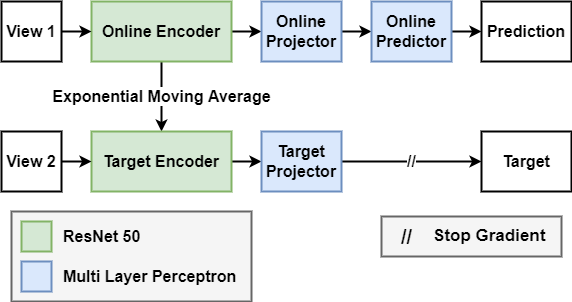}
    \caption{Architecture of sketch-BYOL. It has two encoders based on a ResNet50 model, two projectors and a predictor, each having a fully-connected and a regularization layer with a ReLU activation function. The lower branch produces a target vector for the upper branch to predict.}
    \label{fig:byol}
\end{figure}

We follow the same architecture from BYOL \cite{NEURIPS2020_f3ada80d}, depicted in Figure \ref{fig:byol}. 
It has a ResNet50 for both encoders and an MLP, consisting of a fully connected and a regularization layer with a ReLU activation. 
The model receives $224\times 224$ sketches whose values range between 0 and 255. 
Both the online network and the target network are initialized with weights pre-trained on ImageNet \cite{10.1145/3065386}. We use the original squared $L2$ norm between the prediction and the target vectors.

 This model receives two different views or transformations of the same sketch on each branch. 
 In the image domain, researchers identify 
 six transformations with different selection probabilities. These include color variations and Gaussian filtering. 
However, these transformations do not make sense in a sketch context. Thus,  we propose a set of four specific transformations for sketches, which were selected via ablations studies.  These transformations are:
 
 \begin{itemize}
     \item \textbf{Random Line Skip (probability 0.5):} We randomly delete 10\% of the lines in the sketch.
     \item \textbf{Random Rotation (probability 0.5):} We randomly rotate the sketch with an angle between -30 to 30 degrees.
     \item \textbf{Random Horizontal Flip (probability 0.5):} We randomly flip the sketch horizontally.
     \item \textbf{Random Sized Crop (probability 1.0):} We make a squared crop of the sketch in a random position, with the size of the side being also random between 0.3 and 1.0 times the size of the original sketch.
 \end{itemize}

\subsection{Semi-supervised approaches}

\subsubsection{M2 Semi-supervised model}

\begin{figure}[ht!]
    \centering
    \includegraphics[scale=0.33]{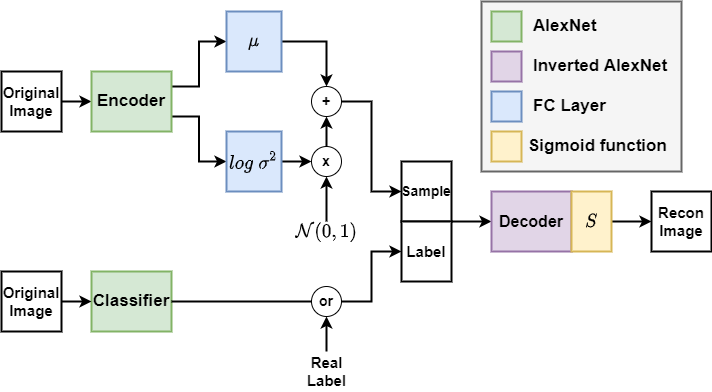}
    \caption{Proposed architecture for M2. Both the encoder and the classifier are AlexNet models, the vectors from the encoder are extracted with fully-connected layers. The decoder is an Inverted AlexNet, it receives the output of the encoder and the classifier to reconstruct the sketch, it also has a sigmoid activation function due to the binary values of a sketch.}
    \label{fig:m2}
\end{figure}

The proposed architecture is shown in Figure \ref{fig:m2}, 
it utilizes an AlexNet backbone for both the encoder and the classifier, and an inverted AlexNet model 
for the decoder. Here, we choose AlexNet over ResNet50 because the last showed signs of underfitting in this scenario. The latent space consists of the concatenation of $\mu$ (32D)  with the classification vector (128D). We represent sketches with value 0 for strokes, and value 1 for background, thus we add a sigmoid function at the end.    

To train our model we use a traditional generative loss $\mathcal{L}$, identically as in VAE. However, in this semi-supervised context, we take advantage of the two worlds, the labeled and unlabeled data, thus we propose two losses, the labeled loss $l\mathcal{L}_{M2}$ and the unlabeled one $u\mathcal{L}_{M2}$ that are defined in Equations \ref{eq:labeled_semi} and \ref{eq:unlabeled_semi}.

\begin{equation}
    l\mathcal{L}_{M2} = \mathcal{L} + 0.1 N\cdot CE(y_{true}, y_{pred})    
    \label{eq:labeled_semi}
\end{equation}
    
\begin{equation}
    u\mathcal{L}_{M2} = \sum y_{pred} \cdot \mathcal{L} +  \mathcal{H}(y_pred)
    \label{eq:unlabeled_semi}
\end{equation}

where $N$ is the length of the training dataset, and $\mathcal{H}$ is the entropy of predictions. In addition, for $u\mathcal{L}_{M2}$, $\mathcal{L}$ is weighted by the confidence of the predictions. 

\subsubsection{Semi-supervised Variational Autoencoder}

\begin{figure}[ht!]
    \centering
    \includegraphics[width=\linewidth]{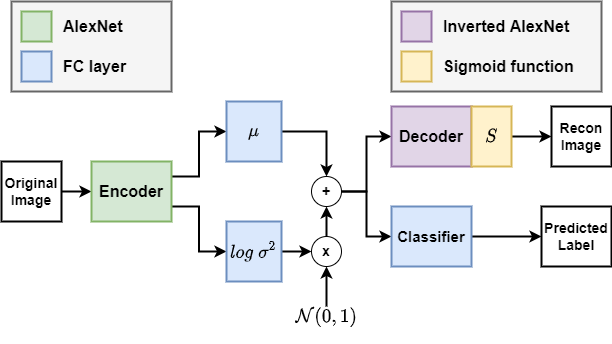}
    \caption{Proposed architecture for the semi-supervised VAE. The encoder consists of an AlexNet model and the latent space vectors are extracted with fully-connected layers. The decoder is an Inverted AlexNet with a sigmoid activation function due to the binary values of a sketch, while the classifier consists of a single fully-connected layer with a softmax activation.}
    \label{fig:ssvae}
\end{figure}

The proposed architecture, shown in Figure \ref{fig:ssvae}, utilizes an AlexNet backbone for the encoder, an inverted AlexNet model 
for the decoder, for the same reasons given for the M2 model, and a single fully-connected layer with a softmax activation function as the classifier. We choose a latent space of 32 dimensions. 

Unlike the M2 model, the output of the classifier is not used as part of the feature vector, we only use $\mu$. 
Like with previous VAE models, we used sketches with dimensions of $256\times 256$, and binary representations for sketches. 
We use a sigmoid activation layer in the output of the decoder, binary cross-entropy as a reconstruction loss, and cross-entropy as the classification loss. 
We used a $\beta$ weight for the KLD loss of $0.1$ and an $\alpha$ weight for the classifier loss of $0.1$\footnote{Selected from preliminary experiments}.

\section{Experiments}
\subsection{Datasets}
For the experiments, we use sketches from the \textit{The Quick, Draw! dataset}\footnote{https://quickdraw.withgoogle.com/data}, a collection of 50 million drawings with 345 classes. To evaluate the effect of using unlabeled data in our models, we define four training configurations
with different percentages of labeled data, as shown in the Table \ref{table:training_datasets}. 
\begin{table}[ht!]
    \centering
    \resizebox{0.99\columnwidth}{!}
    {
        \begin{tabular}{|l|c|c|c|}
            \hline
            Name & \makecell{\% labels} & \makecell{\#classes} & \makecell{samples/class} \\\hline 
            Unlabeled QD  & 0\% & 128 & 1000 \\
            \hline
            10\% labeled QD & 10\% & 128 & 1000 \\
            \hline
            50\% labeled QD & 50\% & 128 & 1000 \\
            \hline
            Labeled QD & 100\% & 128 & 1000\\ 
            \hline
        \end{tabular}
    }
    \caption{Training datasets built from \textit{The Quick, Draw! Dataset}.}
    \label{table:training_datasets}
\end{table}

For each configuration, we randomly select 128 classes. For testing, we define two sets, each one with 100  instances per class:

\begin{itemize}
    \item \textit{\textbf{Known QD}} with the same 128 classes of the train data.
    \item \textit{\textbf{Unknown QD}} with other 128 classes not contained in the train dataset.
\end{itemize}

\subsection{Evaluation protocol}
Both VAE and BYOL are trained with the \textit{Unlabeled QD training dataset}. For M2 and semi-supervised VAE models, 
we use the \textit{10\%} and \textit{50\% labeled QD training datasets}. 
Then, we evaluate the trained models on our test datasets.
We evaluate sketch retrieval with the \textbf{accuracy} of a $kNN$ classifier ($k=5$), and \textbf{mAP@5}. 
The first metric measures how the classes are distributed in the generated latent space
, while the second measures how relevant are the sketches retrieved for each query.
The latent spaces of VAE and semi-supervised VAE have 32 dimensions. For the M2 model, the latent space is defined by the output of the encoder of size 32 and the output of the classifier of size 128, with 
160 dimensions total. Finally, for \textit{sketch-BYOL}, we use the output of a ResNet 50 as feature vectors (2048D).
\subsection{Quantitative experiments}

\begin{table}[ht!]
    \centering
    \resizebox{0.99\columnwidth}{!}
    {
        \begin{tabular}{|l|c|c|c|}
            \hline
            Model & \makecell{Accuracy} & mAP@5 & Type \\\hline
            VAE & 0.338 & 0.307 & self \\\hline
            M2 10\% labeled & 0.528 & 0.492 & semi \\\hline
            M2 50\% labeled & 0.663 & 0.624 & semi \\\hline
            SSVAE 10\% labeled & 0.490 & 0.449 & semi \\\hline
            SSVAE 50\% labeled & 0.672 & 0.648 & semi \\\hline
            sketch-BYOL & 0.634 & 0.597 & self \\\hline
            ResNet & \textbf{0.687} & \textbf{0.655} & supervised \\\hline
        \end{tabular}
    }
    \caption{
    Comparisons on Known QD testing dataset, which contains the same classes as the training datasets. Best results are highlighted in bold. Accuracy is computed from a kNN classifier with k=5.}
    \label{table:results_seen_classes}
\end{table}

Table \ref{table:results_seen_classes} shows the results with known categories. 
Supervised ResNet outperforms other baselines, and in contrast, VAE obtains the worst results having about half the performance in both metrics. In the semi-supervised scenario, adding a percentage of labels in the training datasets improves the results. We observed adding 50\% of the data makes the models competitive. 
On the other hand, \textit{sketch-BYOL} achieves competitive performance, being only five points below the Resnet in both metrics. It is important to highlight \textit{sketch-BYOL} among the top performers without using any labels and being a self-supervised approach. BYOL can be suitable for real-world datasets with no annotations.
\newline

\begin{table}[ht!]
    \centering
    \resizebox{0.99\columnwidth}{!}
    {
        \begin{tabular}{|l|c|c|c|}
            \hline
            Model & \makecell{Accuracy} & mAP@5 & Type \\\hline
            VAE & 0.330 & 0.310 & self \\\hline
            M2 10\% labeled & 0.291 & 0.267 & semi \\\hline
            M2 50\% labeled & 0.318 & 0.298 & semi \\\hline
            SSVAE 10\% labeled & 0.403 & 0.358 & semi \\\hline
            SSVAE 50\% labeled & 0.422 & 0.390 & semi \\\hline
            sketch-BYOL & \textbf{0.627} & \textbf{0.590} & self \\\hline
            ResNet & 0.575 & 0.528 & supervised \\\hline
        \end{tabular}
    }
    \caption{Comparison on Unknown QD testing dataset, which contains different classes than the training datasets. Best results are highlighted in bold. Accuracy is computed from a kNN classifier with k=5.}
    \label{table:results_unseen_classes}
\end{table}

\begin{figure*}[!ht]
    \centering
    \includegraphics[scale = 0.42]{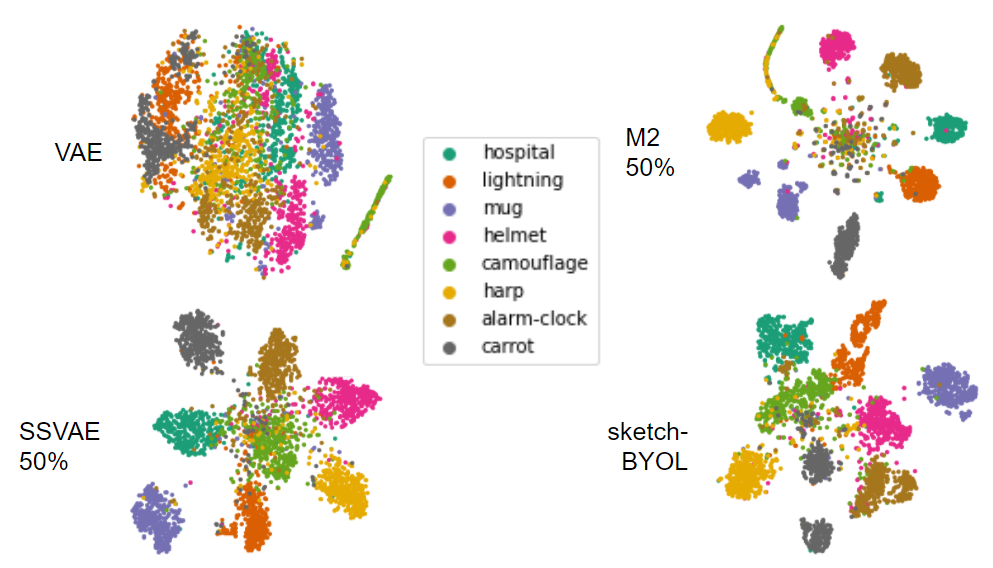} 
    \caption{t-SNE visualization for known categories. Most baselines shows clear group boundaries, with exception of VAE.}
    \label{fig:tsne_seen}
\end{figure*}

When we evaluate with unseen categories (Table \ref{table:results_unseen_classes}), all supervised and semi-supervised models 
decrease their performance metrics. ResNet drops around 10 points in both metrics, while the semi-supervised models suffer an even greater loss, getting closer to VAE. Both self-supervised models don't show a big change in performance, and while VAE keeps being far from competitive, \textit{sketch-BYOL} becomes the best performing model in both metrics.

\subsection{Qualitative experiments}

To understand the behavior of the inferred feature spaces, we visualize the latent space of the studied models together with the class distribution. To this end, we project the real space to 2D by the t-SNE approach \cite{Maaten2008VisualizingDU}. We use a subset of 8 classes, randomly selected, from each evaluation dataset, and observe the differences.  For the semi-supervised models, we only show results with 50\% of labeled data.

In Figure \ref{fig:tsne_seen}, we 
observe the distribution of known categories in the latent space of each model. First, 
VAE learns some clusters with some overlapping categories. For example, the hospital category overlaps with helmet, harp, and, camouflage categories. 
It also seems to occupy the space uniformly, which seems to be the effect of the KLD loss. With the M2 model, we observe well-defined class clusters, but with outliers of all classes in the middle, this might be the effect of concatenating two different vectors to form its latent space. 
Both the semi-supervised VAE and \textit{sketch-BYOL} produce class clusters with very little overlapping, and only the camouflage category seems harder to classify. 
Interestingly, \textit{sketch-BYOL} learns differences within some classes. 
Lightning, alarm clock, and carrot categories show two groups each.

\begin{figure*}[!ht]
    \centering
    \includegraphics[scale = 0.42]{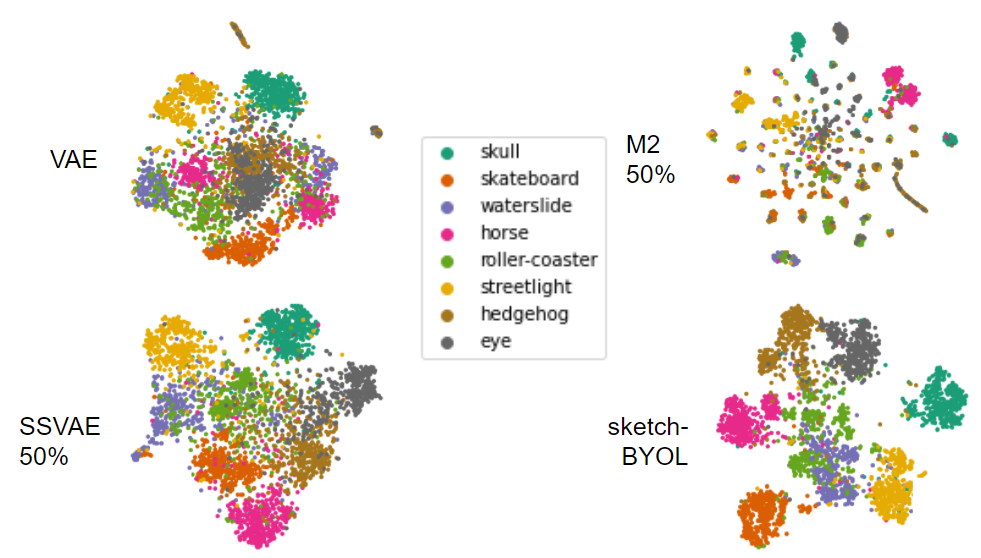}
    \caption{t-SNE visualization for unknown categories. Sketch-BYOL show clear group boundaries as opposed to other baselines.}
    \label{fig:tsne_unseen}
\end{figure*}

When we repeat the protocol evaluation with unknown categories (see Figure \ref{fig:tsne_unseen}), 
VAE presents similar properties as before, forming category clusters 
with overlap categories, 
and instances distributed uniformly in the latent space. On the other hand, both semi-supervised models decrease the quality of their latent spaces. 
For M2, instead of having one cluster per category, it shows multiple smaller clusters with some overlap between them, while the semi-supervised VAE 
has a latent space similar to VAE 
However, \textit{sketch-BYOL} is the only model that still achieves well-defined category clusters, which is consistent with the results from Tables \ref{table:results_seen_classes} and \ref{table:results_unseen_classes}.

\begin{figure*}[!ht]
    \centering
        \includegraphics[scale = 0.32]{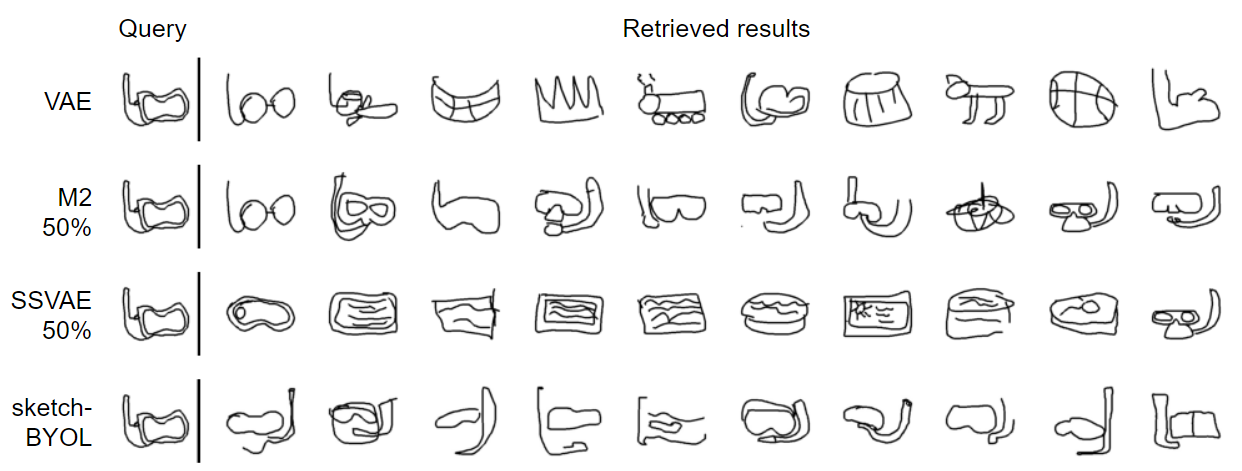}
    \caption{Sample snorkel sketch-retrieval results from known categories.}
    \label{fig:query_seen_1}
\end{figure*}

\begin{figure*}[!ht]
    \centering
    \includegraphics[scale = 0.32]{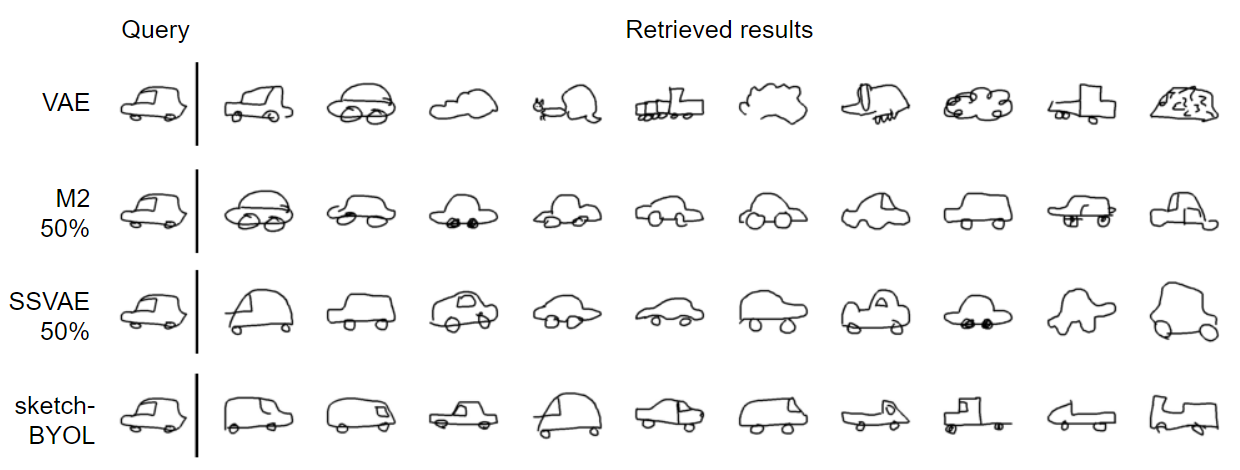}
    \caption{Sample car sketch-retrieval results from known categories.}
    \label{fig:query_seen_2}
\end{figure*}

\begin{figure*}[!ht]
    \centering
    \includegraphics[scale = 0.32]{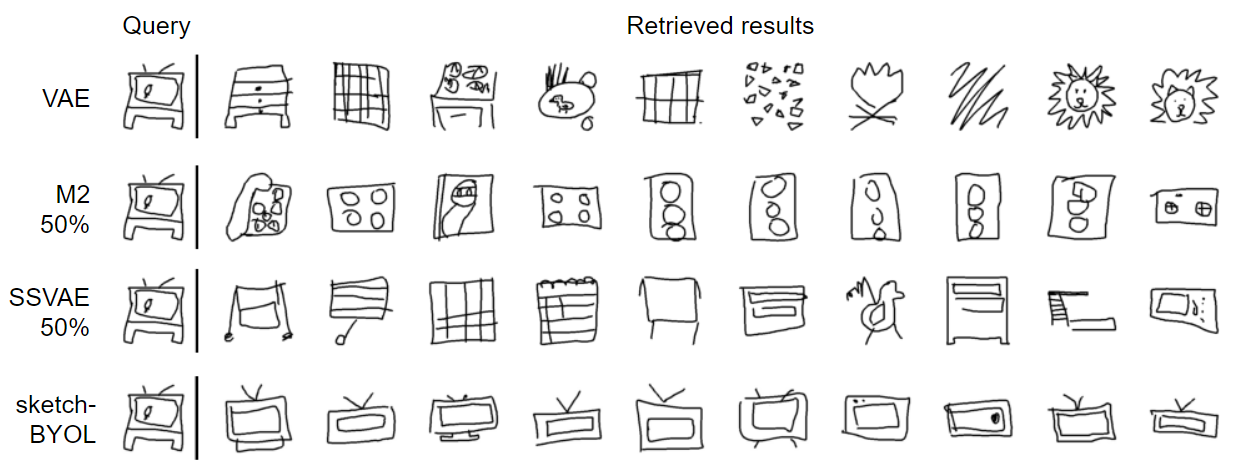}
    \caption{Sample television sketch-retrieval results from unknown categories.}
    \label{fig:query_unseen_1}
\end{figure*}

\begin{figure*}[!ht]
    \centering
    \includegraphics[scale = 0.32]{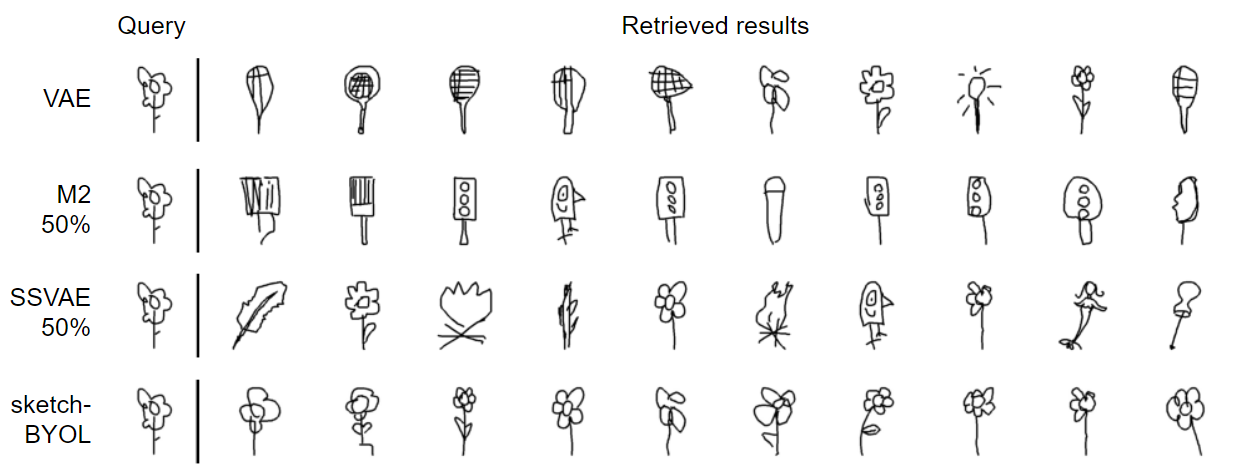}
    \caption{Sample flower sketch-retrieval results from unknown categories.}
    \label{fig:query_unseen_2}
\end{figure*}

\begin{figure*}[!ht]
    \centering
    \includegraphics[scale = 0.32]{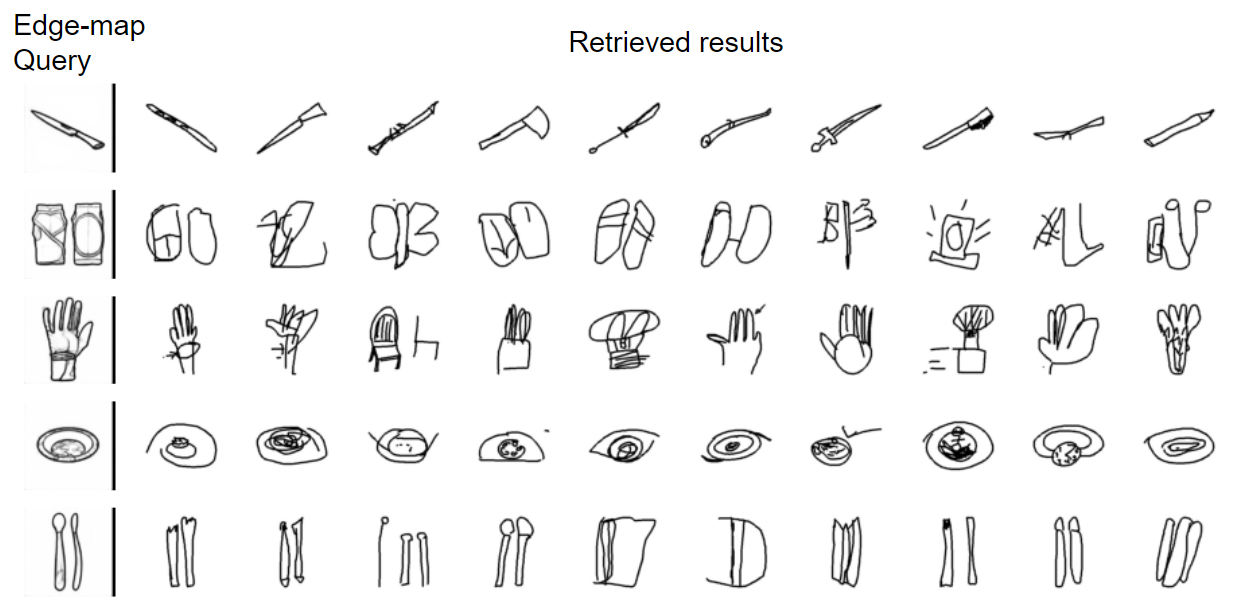}
    \caption{Finding sketches from an image. We convert an image to an edge map using \textit{PiDiNet} \cite{su2021pixel}, and sketch-BYOL retrieves the most similar sketches. 
    }
    \label{fig:application}
\end{figure*}

We also show results for sketch retrieval in Figures \ref{fig:query_seen_1} and \ref{fig:query_seen_2} 
with known categories. In the first figure, for a snorkel query, VAE and semi-supervised VAE are only able to retrieve a few sketches of the correct category, while M2 and \textit{sketch-BYOL} have no trouble with this query preserving fine-grained details such as contours and its breathing tube. In the second figure, for a car query, VAE clearly has the worst performance finding no relevant sketches such as clouds and trains. The remaining models retrieve car sketches with different forms and wheels styles. 

When we evaluate with unknown categories, as shown in Figures \ref{fig:query_unseen_1} and \ref{fig:query_unseen_2}, we see that all VAE-based models have trouble with the queries. In the first figure, for a television query, \textit{sketch-BYOL} finds relevant television results preserving square shape, and TV antenna, while the other models find sketches with very little relevance, confusing with kitchen, drawers, and traffic lights.  In the second figure, for a flower query, \textit{sketch-BYOL} still is the best performer, except this time both VAE and semi-supervised VAE retrieve a few relevant flowers.
\section{Application}

Our sketch-BYOL can be employed to find hand-drawn sketches associated with edgemaps. In this manner, it would possible to produce sketch-image pairs required in tasks like sketch-based image retrieval, sketch-based localization, and sketch2photo translation.  The main advantage is to create more realistic and human-based sketches minimizing human participation, thus allowing self-supervision. Selecting the top k results, we can generate sketch variability to capture different human interpretations and can be naturally used for coarse-grained sketch-based image retrieval. 

Figure \ref{fig:application} shows the retrieval performance of our model in the QuickDraw dataset when the query is an edge-map produced by \textit{PiDiNet} \cite{su2021pixel}. We can regard this strategy as sketch-based data augmentation.
\section{Conclusions}
In this work, we study how to handle unlabeled data to improve sketch-based understanding in terms of retrieval performance. Our sketch-BYOL approach yields competitive results on queries from known categories and is the best performer for queries of unknown categories, showing a better generalization. We also analyze our approach via embeddings visualizations through t-SNE projections. Finally, we show that our proposal is highly relevant in applications like sketch2photo translation or sketch-based image retrieval, where making sketch-image pairs is required. 

In future work, we plan to extend sketch-BYOL to manage sketch-based image retrieval via its two branches (bimodal-BYOL). The former will encapsulate sketches, and the latter will encapsulate images. Then, commonalities in these two modalities are extracted via the loss function.

{\small
\bibliographystyle{ieee_fullname}
\bibliography{references}
}

\end{document}